\pdfoutput=1

\PassOptionsToPackage{prologue,dvipsnames}{xcolor}

\documentclass[11pt]{article}

\usepackage[]{acl}

\usepackage{times}
\usepackage{latexsym}

\usepackage[T1]{fontenc}

\usepackage[utf8]{inputenc}

\usepackage{microtype}

\usepackage{inconsolata}

\usepackage{graphicx}

\usepackage{longtable}

\usepackage{subcaption}

\usepackage{arydshln}

\usepackage{cleveref}

\usepackage{booktabs}

\usepackage{enumitem}

\title{A Fully Automated Pipeline for Conversational Discourse Annotation: \\ Tree Scheme Generation and Labeling with Large Language Models}

\author{Kseniia Petukhova, Ekaterina Kochmar\\
  Mohamed bin Zayed University of Artificial Intelligence \\
  \texttt{\{kseniia.petukhova, ekaterina.kochmar\}@mbzuai.ac.ae}\\}

\begin{document}
  \maketitle
  \begin{abstract}
  Recent advances in Large Language Models (LLMs) have shown promise in automating discourse annotation for conversations. While manually designing tree annotation schemes significantly improves annotation quality for humans and models, their creation remains time-consuming and requires expert knowledge. We propose a fully automated pipeline that uses LLMs to construct such schemes and perform annotation. We evaluate our approach on speech functions (SFs) and the Switchboard-DAMSL (SWBD-DAMSL) taxonomies. Our experiments compare various design choices, and we show that frequency-guided decision trees, paired with an advanced LLM for annotation, can outperform previously manually designed trees and even match or surpass human annotators while significantly reducing the time required for annotation. We release all code and resultant schemes and annotations to facilitate future research on discourse annotation: \url{https://github.com/Kpetyxova/autoTree}.
  \end{abstract}
  
  \section{Introduction}
  \label{s:intro}
  
  Discourse analysis is essential in NLP tasks like dialog management, generation, summarization, and emotion recognition~\citep{liang2020gunrock,chen2021dialogsum,shou2022conversational}. Traditionally, discourse annotation depends on manual expert labeling, which is costly and time-consuming. LLM-based annotation presents a promising alternative, enhancing speed, consistency, and cost-effectiveness~\citep{gilardi2023chatgpt,hao2024fullanno}. However, challenges such as biases and domain limitations necessitate careful prompt design and evaluation.
  
  \begin{figure}[ht]
    \centering
    \includegraphics[width=\linewidth]{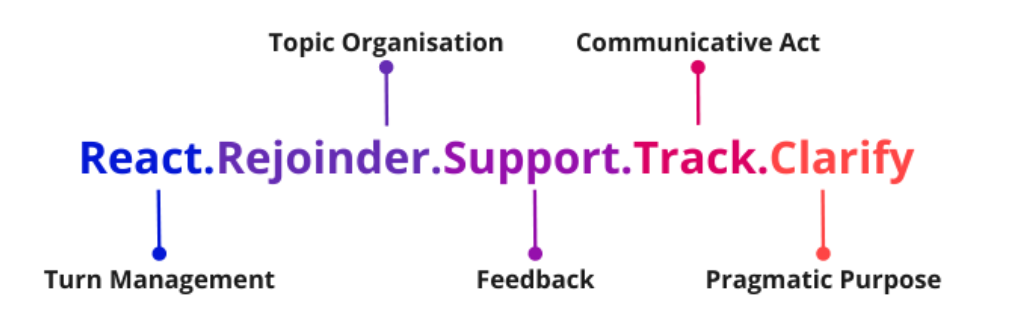}
    \caption{Example of speech function structure~\cite{ostyakova2023chatgpt}.}
    \label{fig:sf}
  \end{figure}
  
  In~\citet{ostyakova2023chatgpt}, the authors explored using \texttt{ChatGPT} to automate discourse annotation for labeling chit-chat dialogs using the speech functions (SFs) taxonomy~\citep{eggins2004analysing}. SFs categorize communicative acts in dialog, capturing speaker intentions and interactions in a hierarchical structure to analyze conversational flow (see Figure~\ref{fig:sf} for an example and Appendix~\ref{ap:sf_taxonomy} for the full label set). \citet{ostyakova2023chatgpt} conducted three sets of experiments: (1) Direct Annotation with an LLM assigning labels from a predefined list of SFs; (2) Step-by-Step Scheme with Intermediate Labels with an LLM selecting labels progressively from broad to specific categories; and (3) Complex Tree-Like Scheme with Yes/No Questions, using a complex tree-like annotation pipeline originally designed for crowdsourced annotation. Since {\em breaking a multi-label selection task into smaller sub-tasks using a tree structure} has improved human performance in complex discourse annotation~\citep{scholman2016step}, the authors hypothesized that the same approach could enhance LLM-based annotation. Prior research also suggests that guiding models with tree-structured prompts significantly improves performance~\citep{yao2024tree}. 
  
  \citet{ostyakova2023chatgpt} found that the Tree-Like Scheme approach enhances LLM accuracy, achieving near-human performance, and suggested that LLMs could serve as a ``silver standard'' for annotation. However, newer and more powerful LLMs have been released since this work was published. These developments not only enhance the potential of LLMs to be used for annotation but also open the door to automating the creation of tree-like schemes,\footnote{We will refer to them as tree schemes.} enabling a fully automated pipeline. This is especially valuable for large taxonomies, such as Intelligent PAL~\citep{morrison2014building} or TEACh-DA~\citep{gella2022dialog} for task-oriented systems, where manually creating tree schemes is complex and time-intensive.
  
  \textbf{This work automates tree schemes creation, making them usable for annotation by crowdsourced workers and LLMs.} A tree scheme is a decision tree that classifies dialog utterances through a series of questions, which can be binary or non-binary, using yes/no or open-ended formats. An example of a tree generated using the pipeline proposed in this work can be found in Appendix~\ref{ap:scheme_example}.

  \section{Related Work}
  \paragraph{Discourse Analysis} 
  
  Researchers analyze discourse structures to improve dialog understanding and management, focusing on pragmatics and speaker intent. One of the key frameworks is the Dialog Act (DA) Theory~\cite{jurafsky1998lexical}, which assigns pragmatic labels to utterances.
  
  The SWBD-DAMSL scheme~\cite{jurafsky1997automatic}, initially created for casual conversations and widely applied to task-oriented systems, classifies dialog acts into 42 classes. Taxonomy of speech functions~\cite{eggins2004analysing} offers a hierarchical annotation approach, integrating DA principles and relational analysis.
  
  Dialog acts are beneficial in task-oriented dialog agents. For example, \citet{gella2022dialog} introduce a scheme for embodied agents, improving natural language interactions and task success, and  \citet{leech2003generic} develop Speech Act Annotated Corpus (SPAAC) scheme, balancing specificity and generalizability in task-oriented dialogs.
  
  \paragraph{LLMs for Discourse and Annotation} 
  
  \citet{ostyakova2023chatgpt} introduce a semi-automated approach for annotating open-domain dialogs using a taxonomy of speech functions and \texttt{ChatGPT}. Their study evaluates three methods: Direct Annotation (selecting from a complete label set), Step-by-Step (progressively narrowing choices), and Tree-Like Schemes (using hierarchical yes/no questions). The Tree-Like Scheme has performed best, particularly for rare classes, achieving high consistency when running the same annotation pipeline with \texttt{ChatGPT} three times (with Fleiss' kappa of 0.83). However, expert input has remained essential for designing the annotation pipelines.
  
  In addition, \citet{yadav2024towards} and \citet{tseng2024expert} explore \texttt{GPT-4}-based semantic annotation, emphasizing prompt design and model limitations. \citet{chen2024large} investigate LLMs for event extraction, addressing data scarcity in fine-tuned models. Finally, \citet{wu2024enhancing} introduce a rationale-driven collaborative framework that refines annotations through iterative reasoning, outperforming standard prompting.

  \section{Speech Functions Corpus}
  
  The experiments described in this work are based on the Speech Functions Corpus, a dataset of dialogs annotated with SFs. This corpus was developed by \citet{ostyakova2023chatgpt}, where three experts, each with at least a B.A. in Linguistics, annotated the DailyDialog dataset~\citep{li2017dailydialog} -- a multi-turn casual dialog dataset -- using the speech functions taxonomy \citep{eggins2004analysing}. The authors of the corpus reduced the original 45 classes proposed by \citet{eggins2004analysing} to a more manageable set of 32 classes.

  The tag set covers five functional dimensions: {\em turn management}, {\em topic organization}, {\em feedback}, {\em communicative acts}, and {\em pragmatic purposes}. While all dimensions are embedded within speech functions, they are distributed unevenly across tags, with individual speech functions incorporating between two and five dimensions. Figure~\ref{fig:sf} illustrates an example of a speech function that includes all dimensions: \textcolor{Blue}{React.}\textcolor{Mahogany}{Rejoinder.}\textcolor{Turquoise}{Support.}\textcolor{BlueGreen}{Track.}\textcolor{BurntOrange}{Clarify}. In this example: (1) \textcolor{Blue}{React} represents {\em turn management}, indicating a speaker change or a reaction to a previous utterance; (2) \textcolor{Mahogany}{Rejoinder} corresponds to {\em topic organization}, signifying active topic development that influences the dialog flow; (3) \textcolor{Turquoise}{Support} denotes {\em feedback}, showing that the speaker is supporting an interlocutor; (4) \textcolor{BlueGreen}{Track} falls under {\em communicative acts}, identifying questions; (5) \textcolor{BurntOrange}{Clarify} serves a {\em pragmatic purpose}, indicating a question aimed at obtaining additional information on the current conversation topic. 
  Some SFs, however, cover fewer dimensions. For instance, \textcolor{Apricot}{Open.}\textcolor{JungleGreen}{Attend} represents only two: {\em turn management} (marking the conversation's beginning) and {\em communicative acts} (a greeting).
  
  The Speech Functions Corpus includes 64 dialogs, containing 1,030 utterances. Appendix~\ref{ap:dialog_example} shows an example of an annotated dialog.
  
  To evaluate the effect of the taxonomy size on the method proposed in this work, the existing taxonomy was converted into the following subsets:\vspace{-0.5em}
  \begin{itemize}
      \item {\bf Full Taxonomy}: This set includes all 32 speech functions labels from \citet{ostyakova2023chatgpt}. A complete list of labels, along with their descriptions, examples, and frequency information, can be found in Appendix~\ref{ap:sf_taxonomy}.\vspace{-0.5em}
      \item {\bf The Top Level of the Taxonomy}: This subset consists of three classes corresponding to the turn management level of the speech functions taxonomy. Definitions for these labels were written manually and are detailed in Appendix~\ref{ap:top_level}. This subset is particularly important because if the model fails to distinguish between these high-level categories, the entire hierarchical structure may be unreliable. \vspace{-0.5em}
      \item {\bf The Top Two Levels of the Taxonomy}: This subset includes six classes, representing a combination of the speech functions taxonomy's turn management and topic organization dimensions. Manual definitions for these classes are provided in Appendix~\ref{ap:two_levels}. The motivation for analyzing this subset is similar to the previous one but with an additional level of complexity, making the classification task more challenging.\vspace{-0.5em}
      \item {\bf Top-20 Frequent Classes of the Taxonomy}: This subset comprises the 20 most frequently occurring classes in the Speech Functions Corpus. It is designed to evaluate how well the model handles frequent classes in the absence of rarer ones.
  \end{itemize}

  \section{Framework for Tree Construction with LLMs}

  Traditionally, discourse annotation has been performed manually by experts or trained annotators, relying on predefined end-class descriptions. However, \citet{scholman2016step} investigates whether non-trained, non-expert annotators can reliably annotate coherence relations using a step-wise approach, which functions similarly to a decision tree. Their findings indicate that a structured step-wise method can indeed make discourse annotation more accessible to non-experts, facilitating large-scale annotation without requiring extensive training. This is achieved by using cognitively plausible primitives rather than relying on complex end labels. A similar observation is made in \citet{ostyakova2023chatgpt}, further supporting the viability of this approach.
  
  To construct an effective decision tree for annotation, it is essential to design questions that do not require expert-level knowledge of discourse but can instead be answered based on the utterance itself. Ideally, related classes should be positioned closely within the tree, forming a hierarchical structure that reflects conceptual similarities between coherence relations. This hierarchical organization not only simplifies decision-making for annotators but also enhances consistency and reliability in annotation.

\begin{figure*}[ht]
    \centering
    \includegraphics[width=\linewidth]{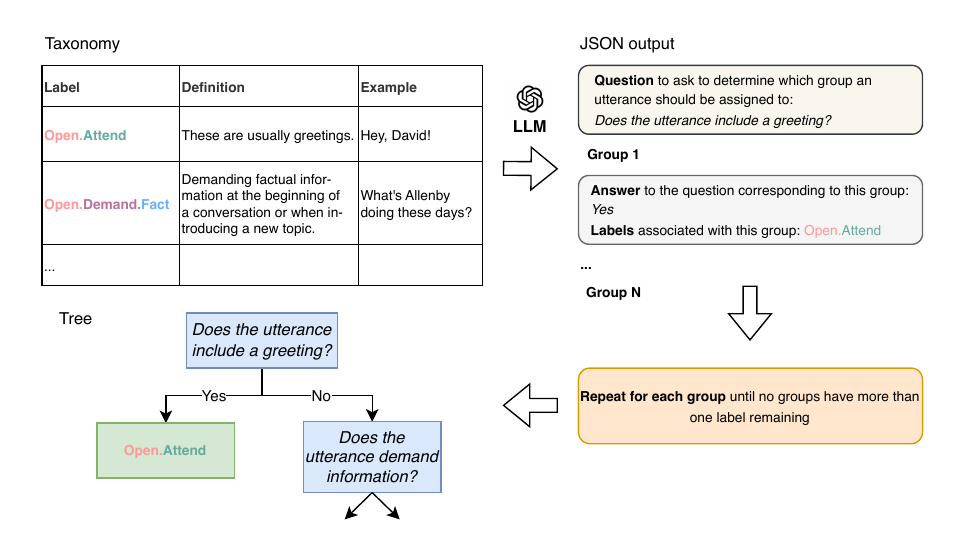}
    \caption{Pipeline for tree construction. An LLM formulates a classification question to split classes into groups, mapping possible answers to respective class groups. This process repeats recursively for created groups until all groups contain only one class. Finally, the grouped data is merged into a single tree structure in JSON format for annotation.}
    \label{fig:pipeline}
  \end{figure*}

  \paragraph{Tree Construction} The pipeline for the tree construction process is illustrated in Figure~\ref{fig:pipeline}. The core concept of the proposed algorithm is to use an advanced LLM to identify distinguishing features that allow it to divide a set of classes into two or more groups. Here, a group refers to a subset of the input classes that the model clusters together based on a shared property it identifies -- framed through a classification question. For example, given the classes \textcolor{Apricot}{Open.}\textcolor{Cerulean}{Demand.}\textcolor{PineGreen}{Fact} (requesting factual information at the beginning of a conversation), \textcolor{Apricot}{Open.}\textcolor{Cerulean}{Demand.}\textcolor{RawSienna}{Opinion} (requesting an opinion at the beginning), \textcolor{Apricot}{Open.}\textcolor{DarkOrchid}{Give.}\textcolor{PineGreen}{Fact} (providing factual information at the beginning), and \textcolor{Apricot}{Open.}\textcolor{DarkOrchid}{Give.}\textcolor{RawSienna}{Opinion} (providing an opinion at the beginning), the model might divide them as \textit{group 1}: \textcolor{Apricot}{Open.}\textcolor{Cerulean}{Demand.}\textcolor{PineGreen}{Fact}, \textcolor{Apricot}{Open.}\textcolor{Cerulean}{Demand.}\textcolor{RawSienna}{Opinion}, and \textit{group 2:} \textcolor{Apricot}{Open.}\textcolor{DarkOrchid}{Give.}\textcolor{PineGreen}{Fact}, \textcolor{Apricot}{Open.}\textcolor{DarkOrchid}{Give.}\textcolor{RawSienna}{Opinion}. To do this, the LLM is provided with a table containing class names, definitions, and usage examples as input.
  
  To enhance reasoning capabilities, we aim for the model to engage in an inner monologue~\citep{zhou2024thinkspeakcultivatingcommunication} before determining how to split the data. To achieve this, the LLM is prompted to first generate a set of questions about an utterance that can aid its understanding. Specifically, the model is instructed to formulate and answer three such reasoning question-answer pairs at the beginning of its response. Recent studies indicate that employing such self-questioning techniques helps models produce more flexible, meaningful outputs, indicating a deeper level of comprehension and reasoning~\citep{sun2024sq}. An example of the LLM's reasoning output can be found in Appendix~\ref{ap:reasoning_example}.
  
  After this reasoning step, the LLM generates a classification question to determine an utterance's group, providing possible answers mapped to class groups. For example, the model might generate a classification question: ``\textit{Is this the beginning of a conversation?}'' The possible answers (groups) could be: (1) ``\textit{Yes, this is the beginning of a conversation.}'' and (2)	``\textit{No, the utterance continues the conversation.}''
  This question helps categorize utterances, splitting the taxonomy into conversation openings and other categories. The prompt template used for this process is detailed in Appendix~\ref{ap:split_into_groups}.

  With taxonomy classes grouped, the process iterates through each group until no group contains more than one class. Once resolved, the resulting JSONs are merged into a single tree JSON for annotation.
  For this step, \texttt{GPT-4} (\texttt{gpt-4-0613}) is used due to its ability to handle long contexts and generate valid JSON outputs~\cite{openai2023gpt4}. The temperature is set to 0.4, which is empirically chosen to balance creativity and reliability, allowing some variation while avoiding excessive randomness. In contrast, a temperature of 0 minimizes variability but may produce overly rigid outputs.

  \paragraph{Annotation} Using the constructed questions tree, dialogs are annotated by traversing the tree for each utterance until reaching a leaf node representing a class label. To ensure a fair comparison with the manually created tree from~\citet{ostyakova2023chatgpt}, the same annotation settings are applied. Specifically, the context length is set to 1, as this yielded the best results in ablation studies. The same model -- \texttt{ChatGPT} (\texttt{\texttt{GPT-3.5}-turbo}) -- is used, along with the optimal temperature of 0.9, as identified in the ablation studies. The prompt format remains consistent with \citet{ostyakova2023chatgpt} and is shown in Appendix~\ref{ap:annotation}. Additionally, open-source models' performance is analyzed in Section~\ref{sec:open_source}.
  
  \paragraph{Evaluation} The framework's performance is assessed by comparing the predicted labels from the annotation step to the gold references provided in~\citet{ostyakova2023chatgpt}, using the same 12 dialogs (189 utterances) as the authors, referred to here as the {\em development set}. To demonstrate generalizability across dialogs, we also evaluate it on another 12 dialogs (165 utterances) from the Speech Functions corpus, referred to as the {\em test set}. To further test generalizability across datasets, we evaluate it on a subset of the Switchboard Dialog Act Corpus — a dataset of telephone conversations annotated using the SWBD-DAMSL annotation scheme, as described in Section~\ref{sec:swda}. The evaluation uses the same metrics as in~\citet{ostyakova2023chatgpt}: Weighted Precision ($P_{w}$), Weighted Recall ($R_{w}$), and Macro F1 ($F1$). Additionally, we report Weighted F1 ($F1_{w}$).
  
  \section{Tree Construction Approaches}
  \label{sec:approaches}
  
  \paragraph{Yes/No Questions vs. Open-Ended Questions} The manually created tree in~\citet{ostyakova2023chatgpt} is binary and composed exclusively of yes/no questions. As the first approach explored in our work, we implement a binary yes/no question tree using the proposed framework. Additionally, we construct a binary tree with open-ended questions. The hypothesis behind this experiment is that allowing the model to generate any questions, rather than limiting it to yes/no questions, can result in a more flexible and nuanced tree for discourse annotation. The prompts used for constructing both the yes/no questions and the open-ended questions tree were developed through iterative refinement. This refinement process involved adjusting and testing prompts until they produced stable and consistent results, ensuring a valid JSON output with all necessary keys present. The final versions of these prompts are provided in Appendix~\ref{ap:prompt_yes_no} and Appendix~\ref{ap:prompt_cats}, respectively.
  
  \paragraph{Binary vs. Non-Binary Trees} Allowing the model to split nodes into multiple groups, rather than restricting it to two groups, has the potential to create a more granular and detailed annotation tree. However, this added granularity may introduce greater complexity and reduce consistency in the annotation process. This experiment is distinct from the previous one, where we only allowed binary open-ended questions and compared them to yes/no questions. As a result, both the questions and answers in that setting were more high-level, distinguishing data based on a single characteristic. In contrast, non-binary open-ended questions allow for greater granularity and specificity, enabling more nuanced differentiation within the data. Experiments described in this section examine the impact of restricting the model to binary splits compared to allowing multiple-group splits on the accuracy of the resulting annotation tree. Both the binary and non-binary trees discussed in this section use open-ended questions. A prompt that allows open-ended questions and splitting data into more than two groups is detailed in Appendix~\ref{ap:prompt_multiple_groups}.
  
  \paragraph{Optimal Split Selection and Backtracking} Inspired by~\citet{yao2024tree}, we evaluate the impact of allowing the model to backtrack while constructing the annotation tree. In each iteration, the model generates three potential splits and assigns a score to each (the prompt used for scoring is provided in Appendix~\ref{ap:prompt_scorer}). These splits are then evaluated using a pre-trained natural language inference (NLI) model,\footnote{\url{https://huggingface.co/cross-encoder/nli-deberta-v3-base}} which classifies them as either contradictory, neutral, or non-contradictory. Among the non-contradictory options, the split with the highest score is selected. If the best-scoring split does not produce a viable partition, the model backtracks and evaluates the next-best option. The motivation behind this experiment is that multiple valid ways to create splits exist, and selecting the seemingly best option at each step may not always result in the most effective tree overall.
  
  \paragraph{Frequency-Guided Optimal Split Selection and Backtracking} Class frequency information can be used to guide the model in making splits and optimize the annotation process. In conversations, certain SFs occur more frequently than others. For instance, \textcolor{Brown}{Sustain.}\textcolor{OrangeRed}{Continue.}\textcolor{Tan}{Prolong.}\textcolor{Maroon}{Extend} appears 21.8\% of the time, representing instances where a speaker adds information to their preceding statement. Similarly,  \textcolor{Blue}{React.}\textcolor{Mahogany}{Rejoinder.}\textcolor{Turquoise}{Support.}\textcolor{BlueGreen}{Track.}\textcolor{BurntOrange}{Clarify} occurs 12\% of the time, typically indicating a question aimed at obtaining additional information. Meanwhile, some SFs are relatively rare. The experiments described in this section aim to construct a decision tree that reflects the distribution of classes, making frequent classes easier and faster to reach compared to rare ones.
  
  To achieve a frequency-guided tree, the prompt used to generate splits is modified as follows: at each step, the model is instructed to create a group containing only the most frequent class if one class is significantly more frequent than the others. The full prompt is provided in Appendix~\ref{ap:prompt_freq}.
  
  \section{Results \& Analysis}
  
  Evaluation results for the Top Level, the Top Two Levels, and Top-20 class subsets on the development set are presented in Tables~\ref{tab:metrics_top_level},~\ref{tab:metrics_two_levels} and~\ref{tab:metrics_top_20} while the results for the complete SF taxonomy on the development and test sets are shown in Tables~\ref{tab:metrics} and~\ref{tab:metrics_test}, respectively.\footnote{The best results for \texttt{GPT-3.5} are highlighted in \textbf{bold}, while the second-best results are marked with an \underline{underline}. The overall best results across all models are highlighted in \textbf{\textcolor{NavyBlue}{blue}}.}
  
  \begin{table}[ht]
      \tiny
      \centering
      \begin{tabular}{lcccc} \hline
           \textbf{Approach} & $\mathbf{P_{w}}$ & $\mathbf{R_{w}}$ & $\mathbf{F1_w}$ & $\mathbf{F1}$ \\ \hline
           Yes/no & 0.55 & 0.34 & 0.37 & 0.33 \\
           Open-ended & \underline{0.68} & \underline{0.63} & \underline{0.61} & \underline{0.63} \\
           Non-binary & \textbf{0.74} & \textbf{0.72} & \textbf{0.70} & \textbf{0.73} \\ \hline
      \end{tabular}
      \caption{Evaluation of annotations on the {\bf development} set using trees constructed through different methods for the \textbf{Top Level} of the SF taxonomy, with \texttt{GPT-3.5} used for annotation.}
      \label{tab:metrics_top_level}
  \end{table}
  
  \begin{table}[ht]
      \centering
      \resizebox{\linewidth}{!}{
      \begin{tabular}{lcccc} \hline
           \textbf{Approach} & $\mathbf{P_{w}}$ & $\mathbf{R_{w}}$ & $\mathbf{F1_w}$ & $\mathbf{F1}$ \\ \hline
           Yes/no & 0.49 & 0.22 & 0.26 & 0.20 \\
           Open-ended & \textbf{0.70} & \textbf{0.65} & \textbf{0.65} & 0.43 \\
           Non-binary & 0.60 & 0.48 & 0.45 & 0.48 \\
           W/ split selection & \underline{0.67} & \underline{0.62} & \underline{0.62} & \textbf{0.60} \\
           Freq.-guided split selection & 0.63 & 0.45 & 0.41 & \underline{0.55} \\ \hdashline
           \parbox[c]{5cm}{W/ split selection\\(\texttt{GPT-4o} for annotation)} & 0.79 & \textbf{\textcolor{NavyBlue}{0.78}} & \textbf{\textcolor{NavyBlue}{0.78}} & \textbf{\textcolor{NavyBlue}{0.80}} \\ \hdashline
           \parbox[c]{5cm}{Freq.-guided split selection\\(\texttt{GPT-4o} for annotation)} & \textbf{\textcolor{NavyBlue}{0.80}} & \textbf{\textcolor{NavyBlue}{0.78}} & \textbf{\textcolor{NavyBlue}{0.78}} & \textbf{\textcolor{NavyBlue}{0.80}} \\ \hline
      \end{tabular}
      }
      \caption{Evaluation of annotations on the {\bf development} set using trees constructed through different methods for the \textbf{Top Two Levels} of the SF taxonomy (with \texttt{GPT-3.5}
  used for annotation unless explicitly stated otherwise).}
      \label{tab:metrics_two_levels}
  \end{table}
  
  \begin{table}[ht]
      \centering
      \resizebox{\linewidth}{!}{
      \begin{tabular}{lcccc} \hline
           \textbf{Approach} & $\mathbf{P_{w}}$ & $\mathbf{R_{w}}$ & $\mathbf{F1_w}$ & $\mathbf{F1}$ \\ \hline
           Yes/no & 0.36 & 0.18 & 0.16 & 0.14 \\
           Open-ended & 0.42 & \underline{0.40} & \underline{0.37} & 0.19 \\
           Non-binary & \underline{0.51} & 0.26 & 0.22 & 0.20 \\
           W/ split selection & 0.39 & 0.36 & 0.34 & \underline{0.27} \\
           Freq.-guided split selection & \textbf{0.62} & \textbf{0.66} & \textbf{0.62} & \textbf{0.35} \\ \hdashline
           \parbox[c]{5cm}{W/ split selection\\(\texttt{GPT-4o} for annotation)} & 0.56 & 0.55 & 0.52 & 0.37 \\ \hdashline
           \parbox[c]{5cm}{Freq.-guided split selection\\(\texttt{GPT-4o} for annotation)} & \textbf{\textcolor{NavyBlue}{0.70}} & \textbf{\textcolor{NavyBlue}{0.69}} & \textbf{\textcolor{NavyBlue}{0.67}} & \textbf{\textcolor{NavyBlue}{0.41}} \\ \hline
      \end{tabular}
      }
      \caption{Evaluation of annotations on the {\bf development} set using trees constructed through different methods for the \textbf{Top-20} classes of the SF taxonomy (with \texttt{GPT-3.5}
  used for annotation unless explicitly stated otherwise).}
      \label{tab:metrics_top_20}
  \end{table}
  
  \begin{table*}[ht]
      \tiny
      \centering
      \resizebox{\linewidth}{!}{
      \begin{tabular}{lcccc} \hline
           \textbf{Approach} & $\mathbf{P_{w}}$ & $\mathbf{R_{w}}$ & $\mathbf{F1_w}$ & $\mathbf{F1}$ \\ \hline
           Manually created tree from~\citet{ostyakova2023chatgpt} (crowdsourced annotation of the full dataset) & 0.71 & 0.60 & - & 0.46 \\ \hline
           Manually created tree from~\citet{ostyakova2023chatgpt} (\texttt{ChatGPT} for annotation) & 0.67 & 0.62 & - & 0.43 \\ \hline
           Yes/no & 0.37 & 0.25 & 0.24 & 0.13 \\
           Open-ended & \underline{0.38} & 0.23 & 0.21 & \textbf{0.23} \\
           Non-binary & \textbf{0.39} & 0.34 & 0.31 & 0.16 \\
           W/ split selection & 0.36 & \underline{0.38} & \textbf{0.35} & 0.16 \\
           Freq.-guided split selection & 0.31 & \textbf{0.43} & \underline{0.34} & \underline{0.19} \\ \hline
           W/ split selection (\texttt{GPT-4o} for annotation) & 0.57 & 0.53 & 0.51 & 0.32 \\
           Freq.-guided split selection (\texttt{GPT-4o} for annotation) & \textbf{\textcolor{NavyBlue}{0.83}} & \textbf{\textcolor{NavyBlue}{0.75}} & \textbf{\textcolor{NavyBlue}{0.74}} & \textbf{\textcolor{NavyBlue}{0.60}} \\ \hline
           W/ split selection (\texttt{Llama-3.1-8B-Instruct} for annotation) & 0.56 & 0.40 & 0.41 & 0.24 \\
           Freq.-guided split selection (\texttt{Llama-3.1-8B-Instruct}  for annotation) & 0.45 & 0.48 & 0.41 & 0.30 \\ \hline
           W/ split selection (\texttt{Mistral-7B-Instruct-v0.3} for annotation) & 0.48 & 0.50 & 0.54 & 0.31 \\
           Freq.-guided split selection (\texttt{Mistral-7B-Instruct-v0.3} for annotation) & 0.33 & 0.44 & 0.32 & 0.18 \\ \hline
      \end{tabular}
      }
      \caption{Evaluation of annotations on the \textbf{development} set (except for the first line) using trees constructed through different methods (with \texttt{GPT-3.5} used for annotation unless explicitly stated otherwise).}
      \label{tab:metrics}
  \end{table*}

  \begin{table}[ht]
      \tiny
      \centering
      \resizebox{\linewidth}{!}{
      \begin{tabular}{lcccc} \hline
           \textbf{Approach} & $\mathbf{P_{w}}$ & $\mathbf{R_{w}}$ & $\mathbf{F1_w}$ & $\mathbf{F1}$ \\ \hline
           Yes/no & 0.20 & 0.21 & 0.17 & 0.12 \\
           Open-ended & \underline{0.35} & 0.22 & 0.21 & 0.16 \\
           Non-binary & 0.31 & \underline{0.31} & \underline{0.27} & 0.16 \\
           W/ split selection & \textbf{0.48} & 0.18 & 0.16 & \underline{0.17} \\
           Freq.-guided split selection & \underline{0.43} & \textbf{0.42} & \textbf{0.37} & \textbf{0.23} \\ \hdashline
           \parbox[c]{2.25cm}{Freq.-guided split selection\\(\texttt{GPT-4o} for annotation)} & \textbf{\textcolor{NavyBlue}{0.77}} & \textbf{\textcolor{NavyBlue}{0.68}} & \textbf{\textcolor{NavyBlue}{0.67}} & \textbf{\textcolor{NavyBlue}{0.46}} \\ \hline
      \end{tabular}
      }
      \caption{Evaluation of annotations on the \textbf{test} set using trees constructed through different methods (with \texttt{GPT-3.5} used for annotation unless explicitly stated otherwise).}
      \label{tab:metrics_test}
  \end{table}

  \paragraph{Yes/No Questions vs. Open-Ended Questions} The findings indicate that \textit{open-ended trees outperform yes/no trees across all metrics and data subsets.} However, for the complete SF taxonomy, both yes/no and open-ended trees perform significantly worse than the manually created tree.
  
  \paragraph{Binary vs. Non-Binary Trees} The findings show that \textit{allowing the model to split data into multiple groups generally outperforms restricting it to binary splits.} This performance difference is more pronounced when the number of classes is smaller. However, as the number of classes increases, the gap narrows, with weighted metrics occasionally favoring the binary approach, specifically showing (1) higher $P_{w}$ and $R_{w}$ for the Top Two Levels (6 classes); (2)  higher $R_{w}$ for the Top-20 classes; (3) higher $P_{w}$ for the complete taxonomy (33 classes) on the test set. Nevertheless, higher macro metrics for non-binary trees suggest improved performance for smaller and less frequent classes. Based on these results, further experiments will allow the model to split data into more than two groups.
  
  \paragraph{Optimal Split Selection and Backtracking} For the Two-Level subset, performance improves compared to the approach that does not use split selection and backtracking. In the Top-20 subset, only the $F1$ metric shows an increase, indicating better performance for less frequent classes. For the complete SF taxonomy, performance on the development set remains comparable to the approach without split selection and backtracking, while on the test set, the $F1$ score is higher.\footnote{Results for the Top Level subset are unavailable, as these trees have only one level.}
  
  The lack of improvement in the complete SF taxonomy stems from the absence of an optimal split at the initial step, causing error propagation throughout the taxonomy. Specific issues include: (1) The model misassigned the \textcolor{Apricot}{Open.}\textcolor{JungleGreen}{Attend} category (which represents greetings at the beginning of a conversation) to the branch ``The utterance involves a request for information,'' and (2) It grouped all \textcolor{Blue}{React} classes under the branch ``The dialog utterance involves a response to a request for information,'' which does not accurately represent them. This misclassification often led the annotation model to misroute utterances to the \textcolor{Brown}{Sustain} branch instead, resulting in unreliable annotations. These issues are not due to the updated approach but rather to the fundamental challenge of generating meaningful splits when dealing with many classes. These errors propagate throughout the taxonomy if the model fails to establish a strong initial split.
  
  \paragraph{Frequency-Guided Optimal Split Selection and Backtracking}
  
  The results indicate that the metrics for the Two-Level subset have decreased compared to the approach without frequency guidance (\Cref{tab:metrics_two_levels}), while they have significantly increased for the Top-20 subset (\Cref{tab:metrics_top_20}). For the complete SF taxonomy, the metrics remained at the same level on the development set (\Cref{tab:metrics}) but improved on the test set (\Cref{tab:metrics_test}).

  Manual analysis revealed that during the annotation step, the model frequently selected incorrect paths, often defaulting to upper-level classes. This behavior was especially prevalent in the most frequent class, \textcolor{Brown}{Sustain.}\textcolor{OrangeRed}{Continue.}\textcolor{Tan}{Prolong.}\textcolor{Maroon}{Extend}.
  Despite this, the tree structure appears logical, and the root question is straightforward: \textit{Does the dialog utterance provide supplementary or contradictory information to the previous statement by the same speaker?} The response options for this question are: (1) ``Dialog utterances that provide supplementary or contradictory information to the previous statement by the same speaker"; (2) ``Dialog utterances that do not provide supplementary or contradictory information to the previous statement by the same speaker".
  While the question is specific, emphasizing conditions about the same speaker and the addition of information, the model often ignored these requirements. In numerous cases, the first response option was incorrectly assigned, even at the start of a conversation.
  
  The distinguishing characteristic of this tree is the heightened granularity and specificity of the questions and labels, with each step designed to determine whether the utterance fits a particular class using a single, targeted question. To assess whether the frequency-guided tree presents too significant a challenge for the \texttt{GPT-3.5} model and whether it might perform better with a more advanced model, \texttt{GPT-4o}~\cite{hurst2024gpt} was used during the annotation step (see Section~\ref{sec:gpt4o}). This approach allowed for a direct comparison between \texttt{GPT-4o} and \texttt{GPT-3.5}. Section~\ref{sec:annotation_gaps} also examines whether the observed differences in metrics are the same for trees created without frequency guidance.
  
  \subsection{\texttt{GPT-4o} for Annotation}
  \label{sec:gpt4o}
  
  Evaluation results comparing annotations by \texttt{GPT-4o} and \texttt{GPT-3.5} on trees constructed using frequency-guided optimal split selection and backtracking for the Two-Level and Top-20 class subsets are presented in Tables~\ref{tab:metrics_two_levels} and~\ref{tab:metrics_top_20}, while Tables~\ref{tab:metrics} and~\ref{tab:metrics_test} show the results for the complete SF taxonomy on the development and test sets, respectively. We note that not only are the differences in metrics highly pronounced, but also \textbf{the $\mathbf{P_{w}}$, $\mathbf{R_{w}}$, and $\mathbf{F1}$ scores for annotations on both the development and test sets for the complete SF taxonomy surpass those obtained for the entire dataset annotated by crowdsourcers in~\citet{ostyakova2023chatgpt}}. This finding underscores that the proposed Frequency-Guided Optimal Split Selection approach, combined with an advanced LLM for annotation, may both outperform manually constructed trees and improve traditional human-driven annotation processes.

  \subsection{Do Annotation Gaps Persist in
  Non-Frequency-Guided Approaches?}
  \label{sec:annotation_gaps}
  
  This section examines whether the substantial differences observed between annotations generated by \texttt{GPT-4o} and \texttt{GPT-3.5} also occur in the non-frequency-guided optimal split selection and backtracking approach.
  
  Tables~\ref{tab:metrics_two_levels},~\ref{tab:metrics_top_20},~\ref{tab:metrics} and~\ref{tab:metrics_test} presents performance metrics for annotations produced using \texttt{GPT-3.5} and \texttt{GPT-4o} with the optimal split selection algorithm, both with and without frequency guidance.
  For the frequency-guided approach, the performance gap between \texttt{GPT-3.5} and \texttt{GPT-4o} becomes more pronounced as the number of classes increases. In contrast, there is no substantial difference in performance for the non-frequency-guided approach when switching from \texttt{GPT-3.5} to \texttt{GPT-4o}. This underscores the clear advantage of the frequency-guided approach when paired with a more advanced model.
  
  \subsection{Error Distribution across Depth Levels}
  
  \Cref{tab:error_depth} decomposes accuracy by the depth at which an annotator model first diverges from the gold path. Several clear trends emerge. First, the use of frequency-guided optimal split selection and backtracking in combination with \texttt{GPT-4o} for annotation nearly eliminates root-level mistakes: the annotator model correctly selects the branch in 98\% of the cases. A similar trend is evident at the second level of the tree, where using \texttt{GPT-4o} with a frequency-guided tree yields 95\% accuracy, significantly outperforming the 68\% accuracy observed with \texttt{GPT-3.5} in the same context.
  
  \begin{table}[t]
  \centering
  \tiny
  \resizebox{\linewidth}{!}{
  \begin{tabular}{@{}clcccccccc@{}}
  \toprule
   & & \multicolumn{2}{c}{\textbf{Split-Sel.}} & \multicolumn{2}{c}{\textbf{Freq-Guided}} \\
  \cmidrule(lr){3-4}\cmidrule(lr){5-6}
  \multicolumn{1}{c}{\textbf{Level}} & \multicolumn{1}{c}{\textbf{Metric}} &
  \multicolumn{1}{c}{\texttt{GPT-3.5}} & \multicolumn{1}{c}{\texttt{GPT-4o}} &
  \multicolumn{1}{c}{\texttt{GPT-3.5}} & \multicolumn{1}{c}{\texttt{GPT-4o}} \\
  \midrule
  1 & Acc. & 94.71 & 74.07 & 85.71 & \textbf{98.41} \\
    & Err\% & 21.28 & 55.06 & 25.00 & \textbf{6.38} \\ \hline
  2 & Acc. & 93.85 & 82.14 & 67.74 & \textbf{95.12} \\
    & Err\% & 23.40 & 28.09 & 46.30 & \textbf{17.02} \\ \hline
  3 & Acc. & \textbf{87.35} & 86.73 & 78.10 & 80.13 \\
    & Err\% & 44.68 & \textbf{16.85} & 21.30 & 65.96 \\ \hline
  4 & Acc. & 97.18 & \textbf{100.00} & 72.00 & 98.48 \\
    & Err\% & 4.26 & \textbf{0.00} & 6.48 & 2.13 \\ \hline
  5 & Acc. & 86.36 & \textbf{100.00} & 88.89 & 88.24 \\
    & Err\% & 6.38 & \textbf{0.00} & 0.93 & 8.51 \\ \hline
  6 & Acc. & - & - & 100.00 & 100.00 \\
    & Err\% & - & - & 0.00 & 0.00 \\
  \bottomrule
  \end{tabular}
  }
  \caption{Accuracy and error percentages (relative to the total number of errors) at each tree depth level produced by \texttt{GPT-3.5} and \texttt{GPT-4o}, using trees constructed with frequency-guided and non-frequency-guided split selection algorithms.}
  \label{tab:error_depth}
  \end{table}
  
  For \texttt{GPT-4o} on frequency-guided trees, errors primarily propagate downward, concentrating at the third level of the tree, where 66\% of all errors occur. Most third-level errors arise from misclassifying   \textcolor{Blue}{React.}\textcolor{Mahogany}{Rejoinder.}\textcolor{Turquoise}{Support.}\textcolor{WildStrawberry}{Response.}\textcolor{Aquamarine}{Resolve} (response that provides the information requested in the question) as \textcolor{Blue}{React.}\textcolor{Mulberry}{Respond.}\textcolor{Rhodamine}{Confront.}\textcolor{Emerald}{Reply.}\textcolor{Gray}{Disagree} (negative response). These misclassifications occur because the annotator model confuses the group label ``Utterances that involve a positive or negative response to a previous statement'' with ``Utterances that provide the information requested in the question.'' The semantic overlap between these groups indicates that the issue lies more in the class definitions rather than a serious annotation error. Another frequent source of error involves the vague \textit{Other utterances} group. Utterances of class \textcolor{Blue}{React.}\textcolor{Mulberry}{Respond.}\textcolor{Turquoise}{Support.}\textcolor{NavyBlue}{Register} (e.g., ``\textit{Yeah,}'' ``\textit{Hmm...},'' ``\textit{Right}'') were grouped under \textit{Other utterances} during tree construction but later misannotated as \textcolor{Blue}{React.}\textcolor{Mulberry}{Respond.}\textcolor{Turquoise}{Support.}\textcolor{Orchid}{Reply.}\textcolor{RoyalBlue}{Affirm} (\textit{positive answers or confirmations, e.g., ``Yes''}). Despite explicit instructions to avoid creating \textit{Other} groups during the tree-creation stage, the model occasionally disregarded these instructions and included them, contributing to misclassification. Notably, fewer than 10\% of errors occur beyond the third level, suggesting most challenges arise earlier in the tree.
  
  \subsection{Consistency of Tree Generation}
  
  An important aspect to evaluate is the \textit{consistency} of the proposed framework: if the decision tree is generated multiple times, how similar or different will the resulting trees be? \Cref{tab:metrics_consistency} shows annotation performance on the test set using three trees built with the frequency-guided split selection algorithm. The first and third runs yield nearly identical results, while the second performs slightly worse. Manual inspection confirms that all three trees are broadly similar, with the drop in the second run due to its poor handling of the frequent class -- \textcolor{Blue}{React.}\textcolor{Mulberry}{Respond.}\textcolor{Turquoise}{Support.}\textcolor{Lavender}{Develop.}\textcolor{Maroon}{Extend}. The primary difference between this second tree and the others lies in its structure: it immediately separates the labels {\em Extend}, {\em Enhance}, and {\em Elaborate} into three distinct terminal nodes at a single step. In contrast, the other two trees first group {\em Enhance} and {\em Elaborate} together, distinguishing them from {\em Extend}, and only in the subsequent step split the remaining group into separate nodes. Overall, the framework demonstrates strong consistency across runs.
  
  \begin{table}[htbp]
      \scriptsize
      \centering
      \begin{tabular}{lcccc} \hline
           \textbf{Run} & $\mathbf{P_{w}}$ & $\mathbf{R_{w}}$ & $\mathbf{F1_w}$ & $\mathbf{F1}$ \\ \hline
           1 & 0.77 & 0.68 & 0.67 & 0.46 \\
           2 & 0.66 & 0.64 & 0.62 & 0.41 \\
           3 & 0.76 & 0.66 & 0.68 & 0.43 \\ \hline
      \end{tabular}
      \caption{Evaluation of annotations on the test set using trees generated across three runs with the frequency-guided optimal split selection approach. Annotations were produced by \texttt{GPT-4o}.}
      \label{tab:metrics_consistency}
  \end{table}

  \subsection{Cost Analysis}
  This section provides a cost analysis for creating trees for SFs and annotating data using these trees.
  
  Generating a non-binary tree for the full SF taxonomy with \texttt{GPT-4} costs approximately \$0.40 and takes 2 minutes (max depth: 3). Annotation with \texttt{GPT-4o} costs \$0.24 per dialog, taking 50 seconds.
  
  Using frequency-guided optimal split selection and backtracking, tree creation costs \$5.48 (with \$4.05 for split-candidates and \$1.43 for scoring) and takes 32 minutes. Without optimal split selection and backtracking, a frequency-guided tree costs \$1.83. In this case, the maximum tree depth is around 7. Annotation costs approximately \$0.36 per dialog and takes about 35 seconds.
  
  For comparison, \texttt{GPT-3.5} annotation with a manually created tree costs \$0.03–\$0.07 per dialog~\cite{ostyakova2023chatgpt}. Crowdsourced annotation costs \$0.12–\$0.22 per dialog, averaging 29 minutes per annotation. While the authors do not specify the time required for tree creation, assuming it exceeds half an hour is reasonable.
  
  Based on these estimates, annotating the entire dataset (64 dialogs) using only human resources would cost approximately \$10.88 and take 31 hours plus additional time for tree creation. In contrast, our best approach, frequency-guided optimal split selection with \texttt{GPT-4o}, would cost around \$20.84 but reduce the total time to approximately 1 hour and 25 minutes, offering significant efficiency and quality benefits despite the higher cost.

  \section{Open-Source Models for Annotation}
  \label{sec:open_source}
  
  Table~\ref{tab:metrics} also presents the results of using two open-source models, \texttt{Mistral-7B-Instruct-v0.3}\footnote{\url{https://huggingface.co/mistralai/Mistral-7B-Instruct-v0.3}} and \texttt{Llama-3.1-8B-Instruct},\footnote{\url{https://huggingface.co/meta-llama/Llama-3.1-8B-Instruct}} compared to the closed-source models \texttt{GPT-3.5} and \texttt{GPT-4o} for the annotation steps with two approaches, frequency-guided and non-frequency-guided optimal split selection with backtracking, which were selected based on their strong performance with other models.
  
  These results indicate that despite having fewer parameters than the closed-source models, both open-source models notably outperform \texttt{GPT-3.5}. Specifically, when using the non-frequency-guided approach, both open-source models achieve performance close to \texttt{GPT-4o} while markedly surpassing \texttt{GPT-3.5}. However, with the frequency-guided approach, \texttt{Llama} continues to outperform \texttt{GPT-3.5} noticeably, but both models' performance metrics fall short of \texttt{GPT-4o}. This trend underscores the substantial performance gains achieved by combining the frequency-guided approach with \texttt{GPT-4o}, as discussed in Section~\ref{sec:annotation_gaps}.

  \section{Evaluation on Switchboard Dialog Act Corpus}
  \label{sec:swda}
  To assess the generalizability of our approach, we evaluate two configurations on the SWBD-DAMSL annotation scheme (42 classes). The configurations are: (1) open-ended questions with a non-binary tree and (2) frequency-guided optimal split selection with backtracking. These configurations were selected as they represent the best non-frequency-guided and frequency-guided approaches.  We use the resulting trees to annotate a randomly sampled subset of the Switchboard Dialogue Act Corpus~\cite{Jurafsky-etal:1997,Shriberg-etal:1998,Stolcke-etal:2000},\footnote{GPL-2.0 license} consisting of 260 utterances.\footnote{We initially selected 300 utterances, but 40 were annotated as ``+'', which, in this taxonomy, indicates that the utterance continues the label of the preceding one. Since these cases do not require actual annotation but rather a repetition of the previous label, we excluded them from the analysis.}
  
  Table~\ref{tab:metrics_swda} presents evaluation metrics using \texttt{GPT-4o}. The results confirm that the proposed framework is effective and generalizable across different taxonomies, including large-scale ones like SWBD-DAMSL, with the Frequency-Guided Optimal Split Selection achieving a Weighted F1 score of 0.61.
  
  \begin{table}[ht]
      \tiny
      \centering
      \resizebox{\linewidth}{!}{
      \begin{tabular}{lcccc} \hline
           \textbf{Approach} & $\mathbf{P_{w}}$ & $\mathbf{R_{w}}$ & $\mathbf{F1_w}$ & $\mathbf{F1}$ \\ \hline
           Non-binary & 0.63 & 0.47 & 0.48 & 0.18 \\ \hline
           Freq.-guided split selection & \textbf{\textcolor{NavyBlue}{0.65}} & \textbf{\textcolor{NavyBlue}{0.63}} & \textbf{\textcolor{NavyBlue}{0.61}} & \textbf{\textcolor{NavyBlue}{0.23}} \\ \hline
      \end{tabular}
      }
      \caption{Evaluation of dialog act annotations from the SWBD-DAMSL annotation scheme, generated by \texttt{GPT-4o}, on a randomly selected set of 260 utterances from the Switchboard Dialogue Act Corpus using non-binary open-ended questions tree and a frequency-guided optimal split selection tree.}
      \label{tab:metrics_swda}
  \end{table}
  \vspace{-0.5em}
  
  \section{Conclusions}
  
  We conducted experiments on generating tree schemes for discourse taxonomies using LLMs. This paper proposes a framework that supports the entire pipeline, from tree construction to dialog annotation. Our configuration with frequency-guided tree creation demonstrates that using LLMs for both tree scheme generation and annotation can yield results that surpass manual tree construction and crowdsourced annotation while significantly reducing the time required for the entire process.
  
  \section*{Limitations}
  A key limitation of the proposed method is the restricted set of models that can be used. The tree creation process requires an advanced model, and the annotation step also benefits from using a more sophisticated model. Another limitation is that non-frequency-guided configurations still underperform compared to manually created trees. This highlights the importance of class frequency information in achieving optimal performance.
  
  Potential directions for future research, motivated by the current limitations, include: (1) exploring larger open-source models for improved taxonomy generation and annotation; (2) conducting experiments on other domains, such as classroom discourse and task-oriented dialog systems; (3) incorporating human feedback to allow the model to self-correct and improve annotation accuracy; (4) enabling self-refinement of the taxonomy by adapting to new, previously unseen dialog examples; and (5) allowing the annotation step to select multiple candidate branches/labels, followed by a final evaluation step that explicitly compares the utterance against the chosen candidates' class definitions to determine the most suitable class.
  
  \section*{Ethical Considerations}
  Dialog data often contains personal or sensitive information, making it essential to anonymize and handle data securely when applying the proposed approach to individual datasets. This is crucial for protecting privacy rights. Beyond this consideration, we do not anticipate any significant risks associated with this work or the use of the proposed framework.
  
  
  
  \bibliography{custom}
  \bibliographystyle{acl_natbib}
  
  \appendix
  
  \onecolumn
  \section{Taxonomy of Speech Functions}
  \label{ap:sf_taxonomy}
      \centering
      \begin{longtable}{|p{0.45\textwidth}|p{0.25\textwidth}|p{0.2\textwidth}|p{0.1\textwidth}|} \hline
           \textbf{Label} & \textbf{Definition} & \textbf{Example} & \textbf{Frequency (\%)}\\ \hline
           \textcolor{Apricot}{Open.}\textcolor{JungleGreen}{Attend} & These are usually greetings. & {\em Hey, David!} & 1.6 \\ \hline
           \textcolor{Apricot}{Open.}\textcolor{Cerulean}{Demand.}\textcolor{PineGreen}{Fact} & Demanding factual information at the beginning of a conversation or when introducing a new topic. & {\em What's Allenby doing these days?} & 2.7 \\ \hline
           \textcolor{Apricot}{Open.}\textcolor{Cerulean}{Demand.}\textcolor{RawSienna}{Opinion} & Demanding judgment or evaluative information from the interlocutor at the beginning of a conversation or when introducing a new topic. & {\em Do we need Allenby in this conversation?} & 1.1 \\ \hline
           \textcolor{Apricot}{Open.}\textcolor{DarkOrchid}{Give.}\textcolor{PineGreen}{Fact} & Providing factual information at the beginning of a conversation or when introducing a new topic. & {\em You met his sister.} & 1.8 \\ \hline
           \textcolor{Apricot}{Open.}\textcolor{DarkOrchid}{Give.}\textcolor{RawSienna}{Opinion} & Providing judgment or evaluative information at the beginning of a conversation or when introducing a new topic. & {\em This conversation needs Allenby.} & 0.9 \\ \hline
           \textcolor{Apricot}{Open.}\textcolor{Goldenrod}{Command} & Making a request, an invitation or command to start a dialog or discussion of a new topic. & {\em Could you tell me about your wedding?} & 1.1 \\ \hline
           \textcolor{Blue}{React.}\textcolor{Mahogany}{Rejoinder.}\textcolor{Turquoise}{Support.}\textcolor{BlueGreen}{Track.}\textcolor{Bittersweet}{Probe} & Requesting a confirmation of the information necessary to make clear the previous speaker's statement. & {\em Because Roman lives in Denning Road also?} & 1.9 \\ \hline
           \textcolor{Blue}{React.}\textcolor{Mahogany}{Rejoinder.}\textcolor{Turquoise}{Support.}\textcolor{BlueGreen}{Track.}\textcolor{CadetBlue}{Check} & Getting the previous speaker to repeat an element or the entire statement that the speaker has not heard or understood. & {\em Straight into the what?} & 0.9 \\ \hline
           \textcolor{Blue}{React.}\textcolor{Mahogany}{Rejoinder.}\textcolor{Turquoise}{Support.}\textcolor{BlueGreen}{Track.}\textcolor{BurntOrange}{Clarify} & Asking a question to get additional information on the current topic of the conversation. Requesting to clarify the information already mentioned in the dialog. & {\em What, before bridge?} & 12.0 \\ \hline
           \textcolor{Blue}{React.}\textcolor{Mahogany}{Rejoinder.}\textcolor{Turquoise}{Support.}\textcolor{BlueGreen}{Track.}\textcolor{CornflowerBlue}{Confirm} & Asking for a confirmation of the information received. & [David: {\em Well, he rang Roman, he rang Roman a week ago.}]\newline Nick: {\em Did he?} & 1.6 \\ \hline
           \textcolor{Blue}{React.}\textcolor{Mahogany}{Rejoinder.}\textcolor{Turquoise}{Support.}\textcolor{WildStrawberry}{Response.}\textcolor{Aquamarine}{Resolve} & The response provides the information requested in the question. & [Fran: {\em Oh what is it called?}]\newline Brad: {\em PhD in Science.} & 8.7 \\ \hline
           \textcolor{Blue}{React.}\textcolor{Mahogany}{Rejoinder.}\textcolor{Rhodamine}{Confront.}\textcolor{WildStrawberry}{Response.}\textcolor{BlueViolet}{Re-challenge} & Offering an alternative position, often an interrogative sentence. & [David: {\em No, Messi is the best}]\newline Nick: \textit{PAUSE}\newline David: {\em The best is Pele} & 0.2 \\ \hline
           \textcolor{Blue}{React.}\textcolor{Mahogany}{Rejoinder.}\textcolor{Rhodamine}{Confront.}\textcolor{Salmon}{Challenge.}\textcolor{Cyan}{Rebound} & Questioning the relevance or reliability of the previous statement, often an interrogative sentence. & [David: {\em This conversation needs Allenby.}]\newline Fay: {\em Oh he's in London. So what can we do?} & 0.5 \\ \hline
           \textcolor{Blue}{React.}\textcolor{Mahogany}{Rejoinder.}\textcolor{Rhodamine}{Confront.}\textcolor{Salmon}{Challenge.}\textcolor{OliveGreen}{Detach} & Terminating the dialogue. & {\em So stick that!} & 0.5 \\ \hline
           \textcolor{Blue}{React.}\textcolor{Mahogany}{Rejoinder.}\textcolor{Rhodamine}{Confront.}\textcolor{Salmon}{Challenge.}\textcolor{SpringGreen}{Counter} & Dismissing the addressee's right to his/her position. & {\em You don't understand, Nick.} & 1.2 \\ \hline
           \textcolor{Blue}{React.}\textcolor{Mahogany}{Rejoinder.}\textcolor{Rhodamine}{Confront.}\textcolor{WildStrawberry}{Response.}\textcolor{Peach}{Refute} & Rejecting a transition to a new topic. & [David: {\em I'm out.}]\newline Fay: {\em You can't do that, it's my birthday.} & 0.1 \\ \hline
           \textcolor{Blue}{React.}\textcolor{Mulberry}{Respond.}\textcolor{Turquoise}{Support.}\textcolor{NavyBlue}{Register} & A manifestation of emotions or a display of attention to the interlocutor. & {\em Yeah. Right. Hmm...} & 6.0 \\ \hline
           \textcolor{Blue}{React.}\textcolor{Mulberry}{Respond.}\textcolor{Turquoise}{Support.}\textcolor{Melon}{Engage} & Drawing attention or a response to a greeting. & {\em Hey! Hi-hi.} & 0.6 \\ \hline
           \textcolor{Blue}{React.}\textcolor{Mulberry}{Respond.}\textcolor{Turquoise}{Support.}\textcolor{Orchid}{Reply.}\textcolor{Plum}{Accept} & Expressing gratitude. & {\em Thank you!} & 1.2 \\ \hline
           \textcolor{Blue}{React.}\textcolor{Mulberry}{Respond.}\textcolor{Turquoise}{Support.}\textcolor{Orchid}{Reply.}\textcolor{RoyalBlue}{Affirm} & A positive answer to a question or confirmation of the information provided. Yes/its synonyms or affirmation. & [Nick: {\em He went to London.}]\newline Fay: {\em He did.} & 3.7 \\ \hline
           \textcolor{Blue}{React.}\textcolor{Mulberry}{Respond.}\textcolor{Turquoise}{Support.}\textcolor{Orchid}{Reply.}\textcolor{Red}{Acknowledge} & Indicating knowledge or understanding of the information provided. & {\em I know. I see. Oh yea.} & 1.1 \\ \hline
           \textcolor{Blue}{React.}\textcolor{Mulberry}{Respond.}\textcolor{Turquoise}{Support.}\textcolor{Orchid}{Reply.}\textcolor{SeaGreen}{Agree} & Agreement with the information provided. In most cases, the information that the speaker agrees with is new to him. & {\em Yes. Right.} & 3.8 \\ \hline
           \textcolor{Blue}{React.}\textcolor{Mulberry}{Respond.}\textcolor{Turquoise}{Support.}\textcolor{Lavender}{Develop.}\textcolor{Maroon}{Extend} & Adding supplementary or contradictory information to the previous statement. & David: [{\em That's what the cleaner---your cleaner lady cleaned my place though.}]\newline Nick: {\em She won't come back to our place.} & 8.6 \\ \hline
           \textcolor{Blue}{React.}\textcolor{Mulberry}{Respond.}\textcolor{Turquoise}{Support.}\textcolor{Lavender}{Develop.}\textcolor{Yellow}{Enhance} & Adding details to the previous statement, adding information about time, place, reason, etc. & [Fay: {\em He kept telling me I've got a big operation on with.}]\newline Nick: {\em The trouble with Roman though is that---you know he does still like cleaning up.} & 0.4 \\ \hline
           \textcolor{Blue}{React.}\textcolor{Mulberry}{Respond.}\textcolor{Turquoise}{Support.}\textcolor{Lavender}{Develop.}\textcolor{Violet}{Elaborate} & Clarifying/rephrasing the previous statement or giving examples to it. A declarative sentence or phrase (may include \textit{for example}, \textit{I mean}, \textit{like}). & [Nick: {\em Cause all you'd get is him bloody raving on.}]\newline Fay: {\em He's a bridge player, a naughty bridge player.} & 0.2 \\ \hline
           \textcolor{Blue}{React.}\textcolor{Mulberry}{Respond.}\textcolor{Rhodamine}{Confront.}\textcolor{Orchid}{Reply.}\textcolor{Green}{Disavow} & Denial of knowledge or understanding of information. & {\em I don't know. No idea.} & 0.4 \\ \hline
           \textcolor{Blue}{React.}\textcolor{Mulberry}{Respond.}\textcolor{Rhodamine}{Confront.}\textcolor{Emerald}{Reply.}\textcolor{Gray}{Disagree} & Negative answer to a question or denial of a statement. No, negative sentence. & [David: {\em Is he in London?}] \newline Nick: {\em No.} & 2.0 \\ \hline
           \textcolor{Blue}{React.}\textcolor{Mulberry}{Respond.}\textcolor{Rhodamine}{Confront.}\textcolor{Orchid}{Reply.}\textcolor{ForestGreen}{Contradict} & Refuting previous information. Sentence with opposite polarity. If the previous sentence is negative, then this sentence is positive, and vice versa. & [Fay: {\em Suppose he gives you a hard time, Nick?}]\newline Nick: {\em Oh I like David a lot.}& 0.4 \\ \hline
           \textcolor{Blue}{React.}\textcolor{Mulberry}{Respond.}\textcolor{Goldenrod}{Command} & Making a request, an invitation, or command in response to previous information. & {\em Could you tell me about your wedding?} & - \\ \hline
           \textcolor{Brown}{Sustain.}\textcolor{OrangeRed}{Continue.}\textcolor{LimeGreen}{Monitor} & Checking the involvement of the listener or trying to pass on the role of speaker to them. & {\em You know? Right?} & 0.2 \\ \hline
           \textcolor{Brown}{Sustain.}\textcolor{OrangeRed}{Continue.}\textcolor{Goldenrod}{Command} & Making a request, an invitation, or command to continue the dialog or discussion without changing the speaker. & {\em Could you tell me about your wedding?} & - \\ \hline
           \textcolor{Brown}{Sustain.}\textcolor{OrangeRed}{Continue.}\textcolor{Tan}{Prolong.}\textcolor{Maroon}{Extend} & Adding supplementary or contradictory information to the previous statement. Used only when the speaker remains the same as in the previous utterance. & {\em Just making sure you don't miss the boat. I put it out on Monday mornings. I hear them. I hate trucks.} & 21.8 \\ \hline
           \textcolor{Brown}{Sustain.}\textcolor{OrangeRed}{Continue.}\textcolor{Tan}{Prolong.}\textcolor{Yellow}{Enhance} & Adding details to the previous statement, adding information about time, place, reason, etc. Used only when the speaker remains the same as in the previous utterance. & {\em Nor for much longer. We're too messy for him.} & 5.1 \\ \hline
           \textcolor{Brown}{Sustain.}\textcolor{OrangeRed}{Continue.}\textcolor{Tan}{Prolong.}\textcolor{Violet}{Elaborate} & Clarifying/rephrasing the previous statement or giving examples to it. Used only when the speaker remains the same as in the previous utterance. & {\em Yeah but I don't like people... um... I don't want to be INVOLVED with people.} & 7.9 \\ \hline
           \caption{Taxonomy of speech functions (``-'' indicates that these labels are counted together with \textcolor{Apricot}{Open.}\textcolor{Goldenrod}{Command}.)}
           \label{tab:sf_tax}
      \end{longtable}
  
  \section{Example of Tree Scheme}
  \label{ap:scheme_example}
  \begin{figure}[ht]
    \centering
    \includegraphics[width=\linewidth]{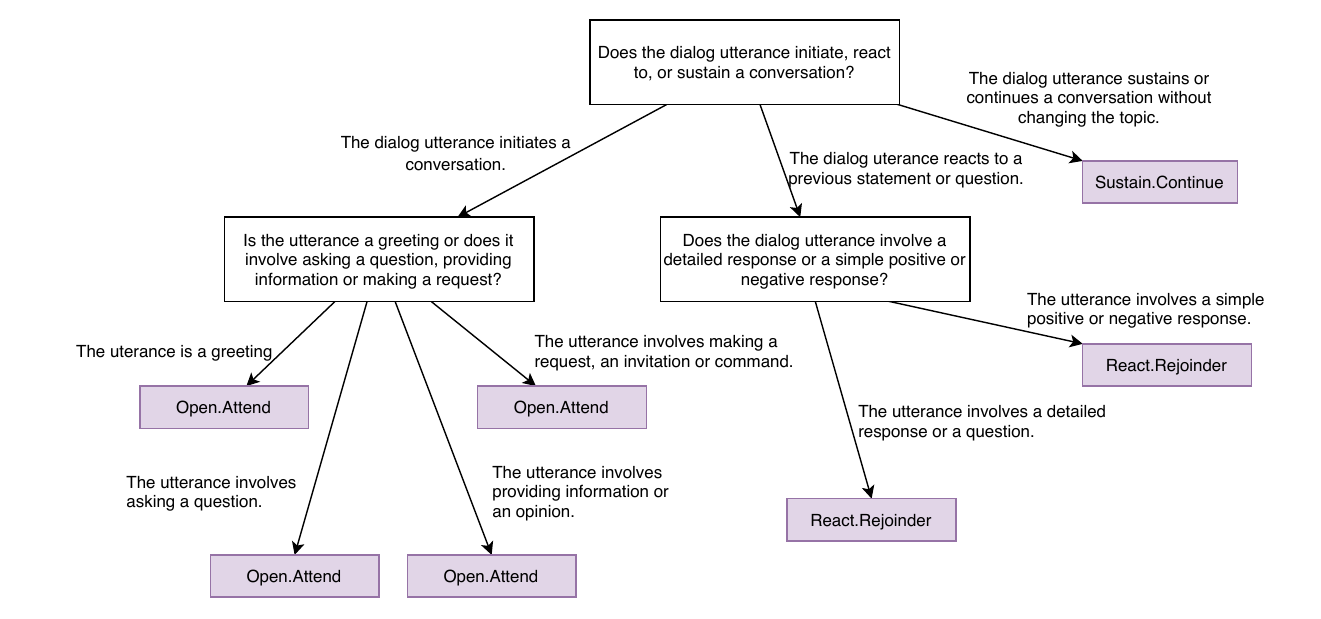}
    \caption{An example of a tree scheme generated by \texttt{GPT-4} using our proposed approach, where nodes represent questions about an utterance, arrows indicate possible answer choices and purple leaves correspond to taxonomy labels.}
    \label{fig:scheme_example}
  \end{figure}

  \newpage
  \section{Example of dialog annotation with Speech Functions}
  \label{ap:dialog_example}
  \begin{figure}[h]
    \centering
    \includegraphics[width=0.5\linewidth]{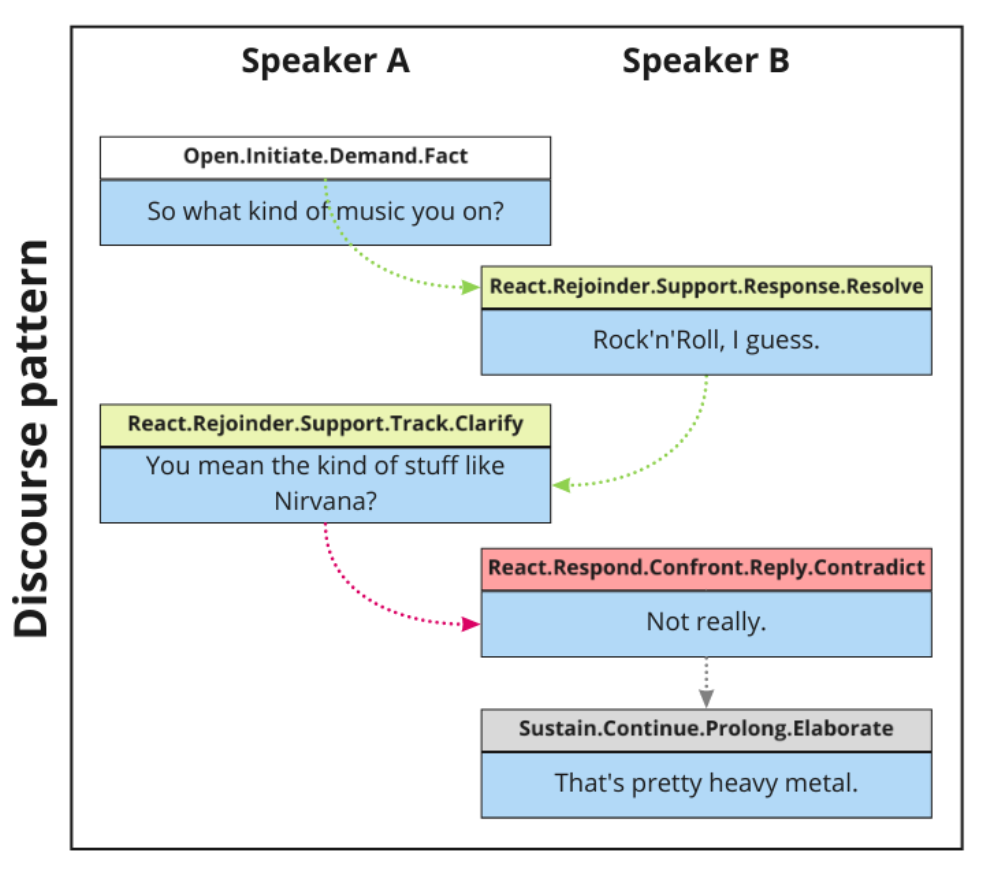}
    \caption{Example of dialog annotation with speech functions~\cite{ostyakova2023chatgpt}.}
    \label{fig:dialog_example}
  \end{figure}
  
  \newpage
  
  \section{The Top Level Labels of SF taxonomy}
  \label{ap:top_level}
  \begin{table*}[h]
      \centering
      \resizebox{\linewidth}{!}{
      \begin{tabular}{|p{0.1\textwidth}|p{0.9\textwidth}|} \hline
           \textbf{Label} & \textbf{Definition} \\ \hline
           \textcolor{Apricot}{Open} & Open utterances are statements or actions that initiate a conversation or introduce a new topic within an ongoing discussion. These may include greetings, questions, requests, invitations, or the sharing of information. \\ \hline
           \textcolor{Blue}{React} & React utterances are responses to the interlocutor’s statements. These may include answers to questions, follow-up questions, emotional reactions, sharing information, expressions of agreement or disagreement, and more. \\ \hline
           \textcolor{Brown}{Sustain} & Sustain utterances are those that extend the speaker’s own preceding statements by adding information, providing new details, or rephrasing. The ``Sustain'' label is applied only when the current and preceding utterances are made by the same speaker. These utterances cannot take the form of questions,  except when the question serves to confirm that the listener is paying attention. \\ \hline
      \end{tabular}
      }
      \caption{Definitions of \textcolor{Apricot}{Open}, \textcolor{Blue}{React} and \textcolor{Brown}{Sustain} labels written manually.}
      \label{tab:definitions_high_levels}
  \end{table*}

  
  \section{The Top Two Level Classes of SF taxonomy}
  \label{ap:two_levels}

  \begin{table*}[h]
      \centering
      \resizebox{\linewidth}{!}{
      \begin{tabular}{|p{0.2\textwidth}|p{0.8\textwidth}|} \hline
           \textbf{Label} & \textbf{Definition}\\ \hline
           \textcolor{Apricot}{Open.}\textcolor{Cerulean}{Demand} & Questions at the beginning of a conversation or when introducing a new topic. \\ \hline
           \textcolor{Apricot}{Open.}\textcolor{DarkOrchid}{Give} & Providing information or opinion at the beginning of a conversation or when introducing a new topic. \\ \hline
           \textcolor{Apricot}{Open.}\textcolor{Goldenrod}{Command} & Making a request, an invitation or command to start a dialog or discussion of a new topic. \\ \hline
           \textcolor{Apricot}{Open.}\textcolor{JungleGreen}{Attend} & These are usually greetings. \\ \hline
           \textcolor{Brown}{Sustain.}\textcolor{OrangeRed}{Continue} & These are used only when there is no change in the speaker from the previous utterance, except for cases when the utterance is a reply to a greeting. The \textcolor{Brown}{Sustain.}\textcolor{OrangeRed}{Continue} class involves adding additional information or details to the speaker's previous statement. It can also include questions intended to check if the interlocutor is listening, as well as requests or invitations. \\ \hline
           \textcolor{Blue}{React.}\textcolor{Mahogany}{Rejoinder} & These include any type of question or detailed response to the interlocutor’s questions, expressions of emotion, and grounding utterances such as ``hmm,'' ``aha,'' and similar reactions. \\ \hline
           \textcolor{Blue}{React.}\textcolor{Mulberry}{Respond} & These include positive or negative responses to questions and expressions of understanding or misunderstanding. They also include the provision of new information or details, similar to \textcolor{Brown}{Sustain.}\textcolor{OrangeRed}{Continue}. However, in this case, the speaker of the current utterance differs from the speaker of the previous one (this rule regarding speakers applies only to cases similar to \textcolor{Brown}{Sustain.}\textcolor{OrangeRed}{Continue}). \\ \hline
      \end{tabular}
      }
      \caption{Definitions of the two highest-level labels.}
      \label{tab:definitions_two_levels}
  \end{table*}

  \newpage
  \section{Example of the Model's Reasoning Output}
  \label{ap:reasoning_example}
  \begin{figure*}[h]
    \centering
    \includegraphics[width=0.8\linewidth]{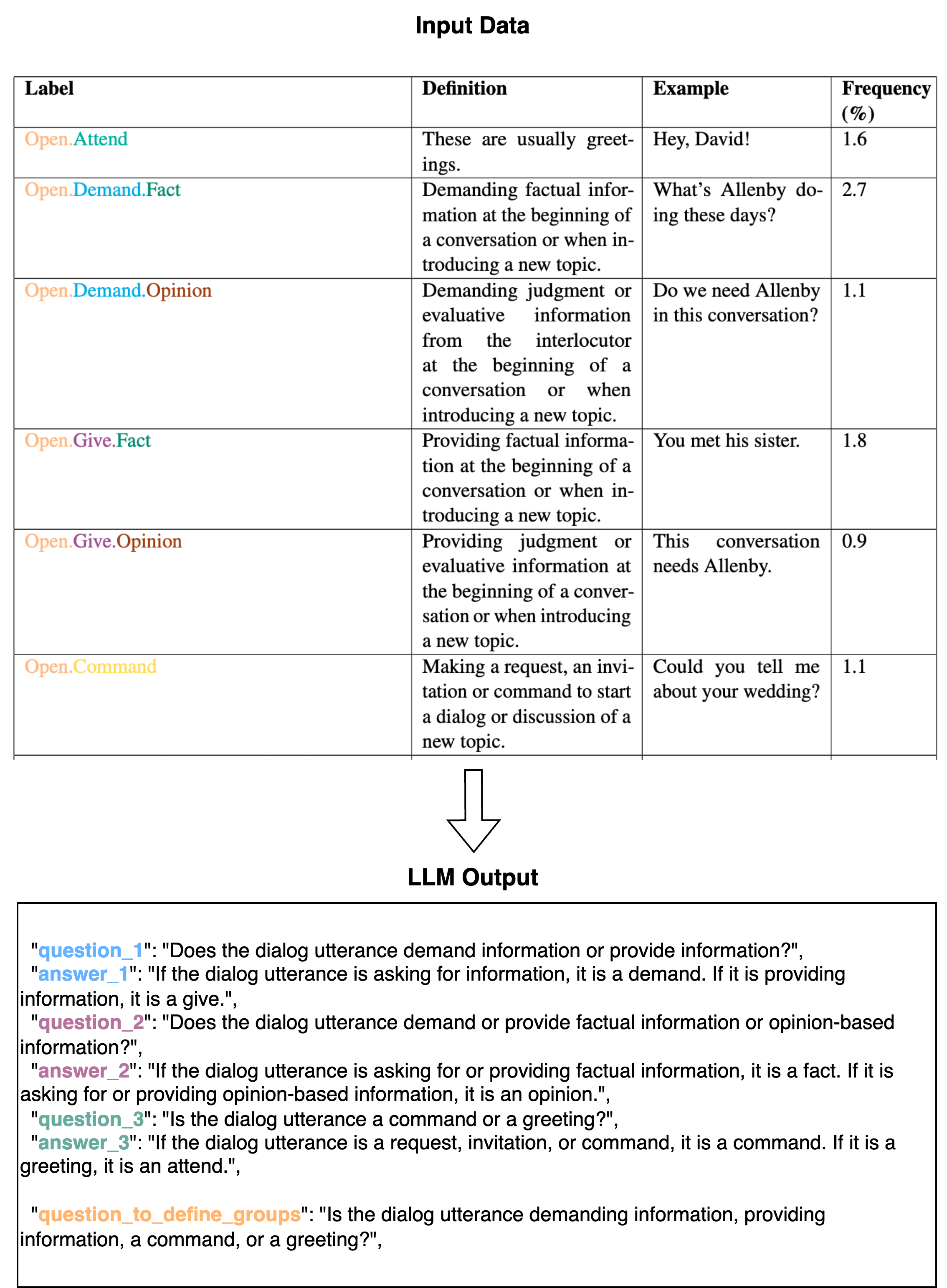}
    \caption{An example of question-answer pairs generated by \texttt{GPT-4} to reason about identifying utterances of different classes.}
    \label{fig:reasoning_example}
  \end{figure*}

  \newpage
  \section{Prompt template for Tree Construction}
  \label{ap:split_into_groups}
  \begin{figure*}[h]
    \centering
    \includegraphics[width=0.7\linewidth]{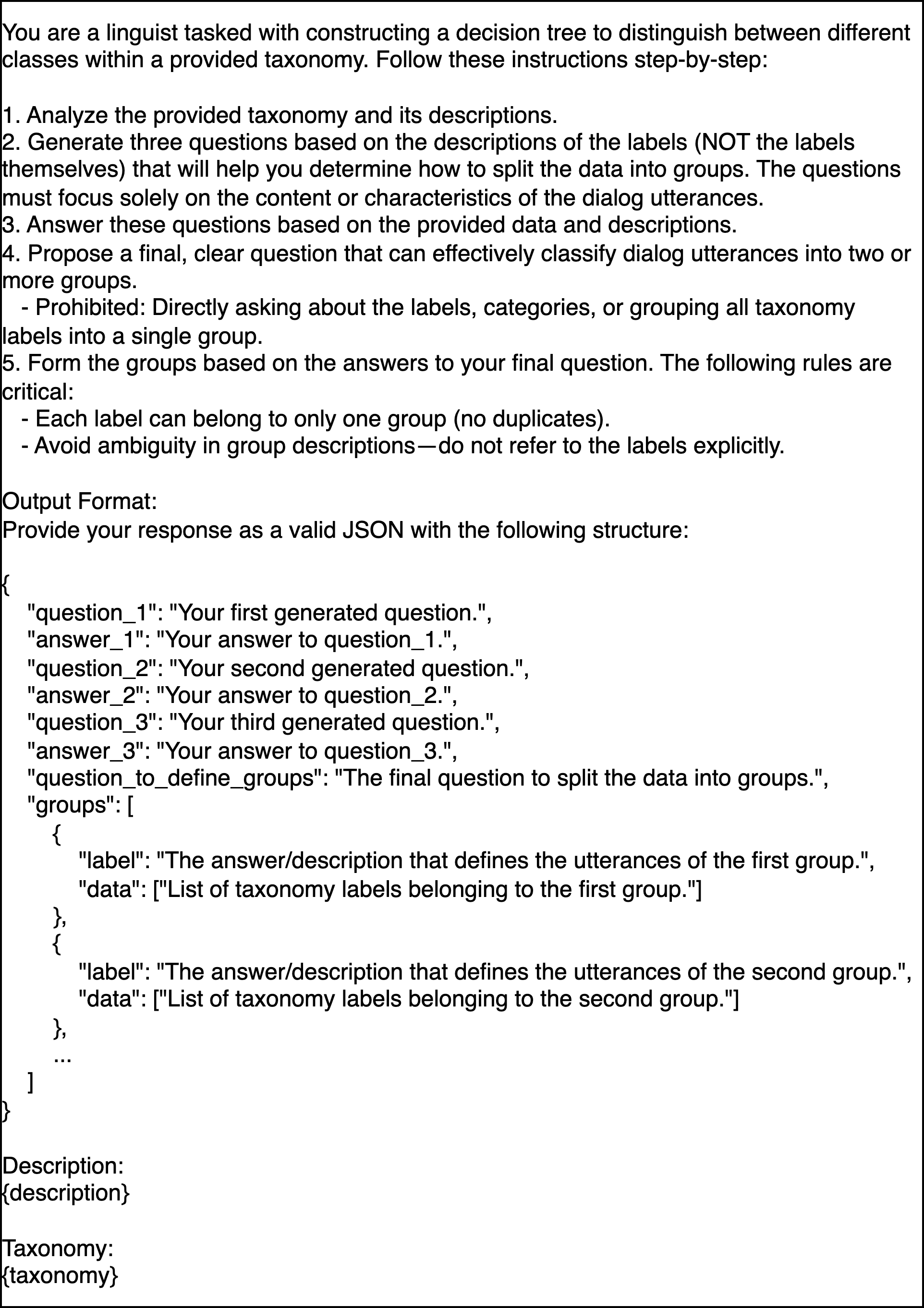}
    \caption{The prompt template used for all experiments with tree construction.}
    \label{fig:split_into_groups}
  \end{figure*}

  \newpage
  \section{Prompt template for the annotation step}
  \label{ap:annotation}
  \begin{figure*}[h]
    \centering
    \includegraphics[width=0.6\linewidth]{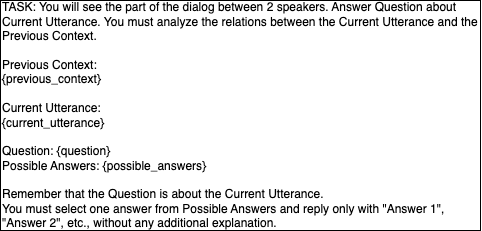}
    \caption{The prompt template used for the annotation step.}
    \label{fig:annotation}
  \end{figure*}

  \newpage
  \section{Prompt for Constructing a Yes/No Questions Tree}
  \label{ap:prompt_yes_no}
  \begin{figure*}[h]
    \centering
    \includegraphics[width=0.8\linewidth]{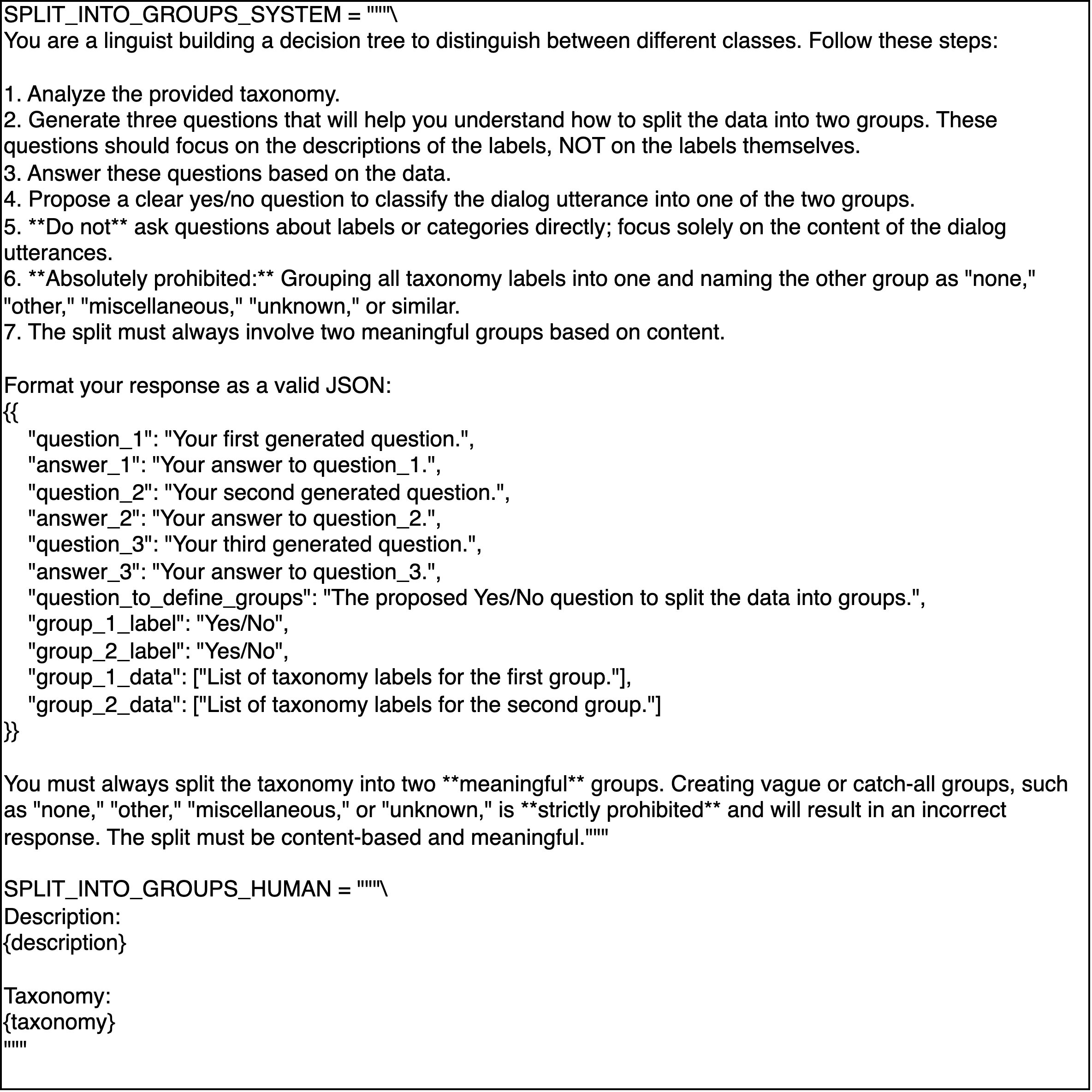}
    \caption{The prompt used for constructing trees with yes/no questions.}
    \label{fig:prompt_yes_no}
  \end{figure*}
  
  \newpage
  \section{Prompt for Constructing an Open-Ended Questions Tree}
  \label{ap:prompt_cats}
  \begin{figure*}[h]
    \centering
    \includegraphics[width=0.8\linewidth]{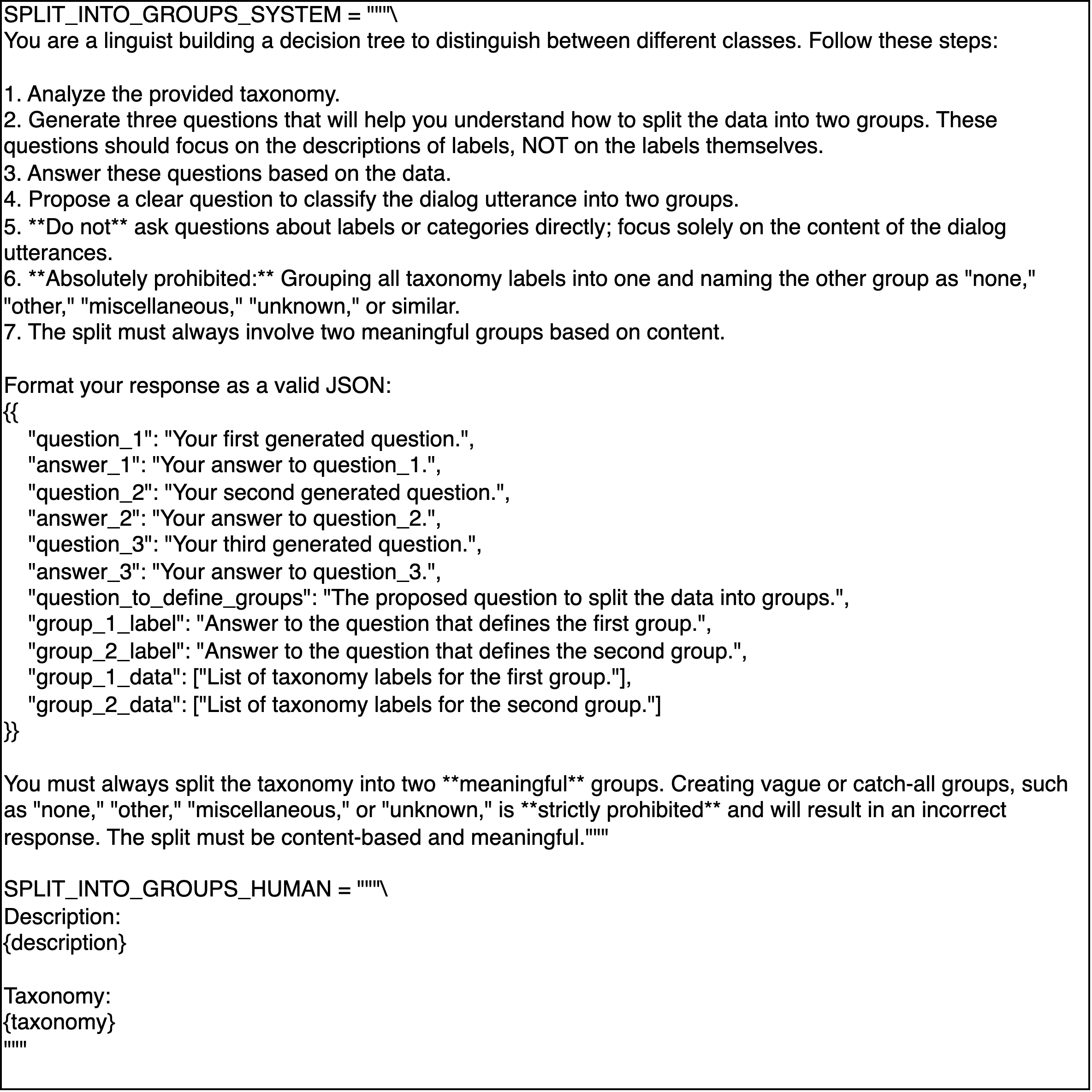}
    \caption{The prompt used for constructing trees with open-ended questions.}
    \label{fig:prompt_cats}
  \end{figure*}
  
  \newpage
  \section{Prompt for Constructing a Non-Binary Open-Ended Questions Tree}
  \label{ap:prompt_multiple_groups}
  \begin{figure*}[h]
    \centering
    \includegraphics[width=0.8\linewidth]{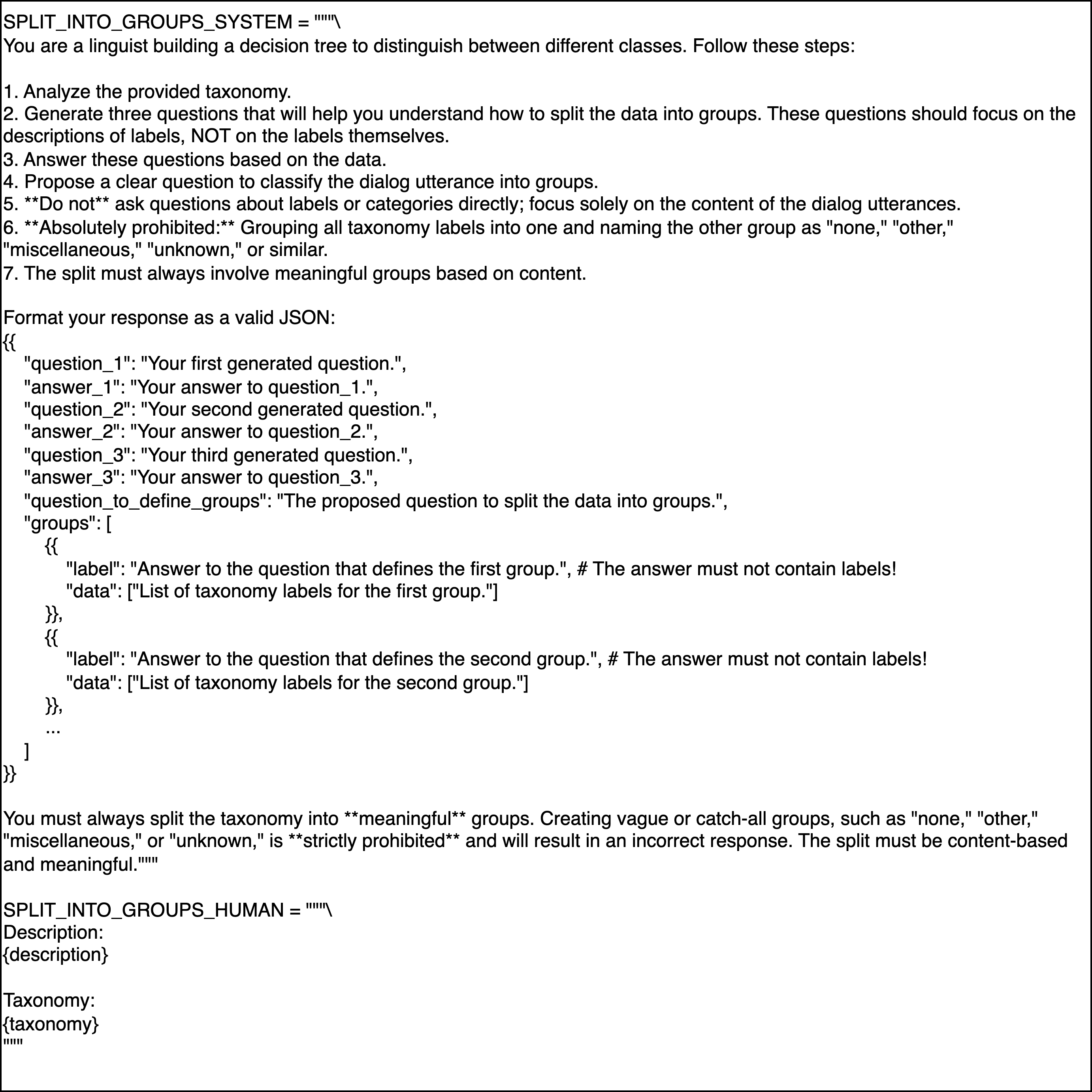}
    \caption{The prompt used for constructing non-binary trees with open-ended questions.}
    \label{fig:prompt_multiple_groups}
  \end{figure*}

  \newpage
  \section{Prompt For Scoring Splits}
  \label{ap:prompt_scorer}
  \begin{figure*}[h]
    \centering
    \includegraphics[width=0.8\linewidth]{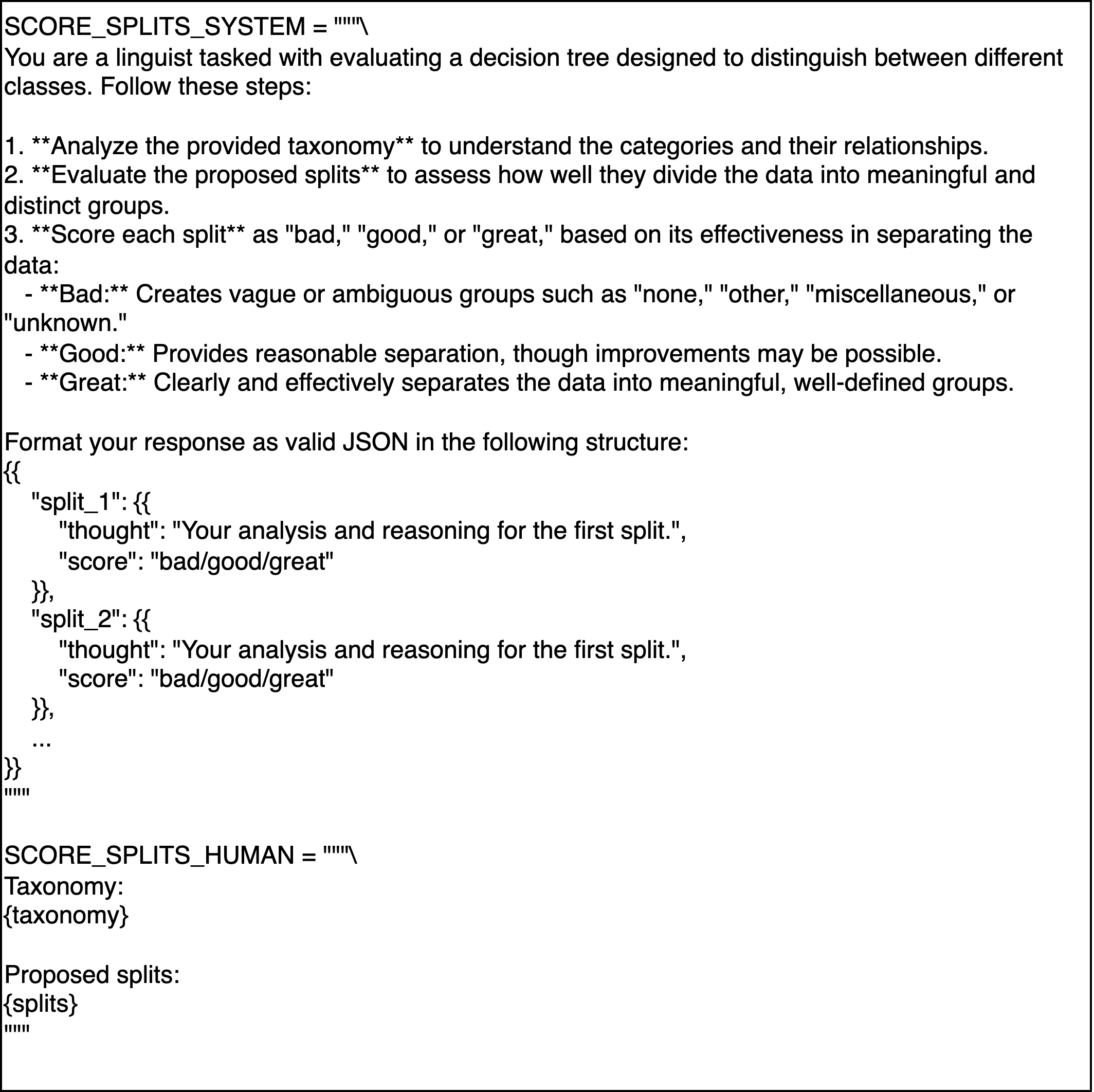}
    \caption{The prompt used to score split-candidates.}
    \label{fig:prompt_scorer}
  \end{figure*}

  \newpage
  \section{Prompt For Frequency-Guided Tree Creation}
  \label{ap:prompt_freq}
  \begin{figure*}[h]
    \centering
    \includegraphics[width=0.8\linewidth]{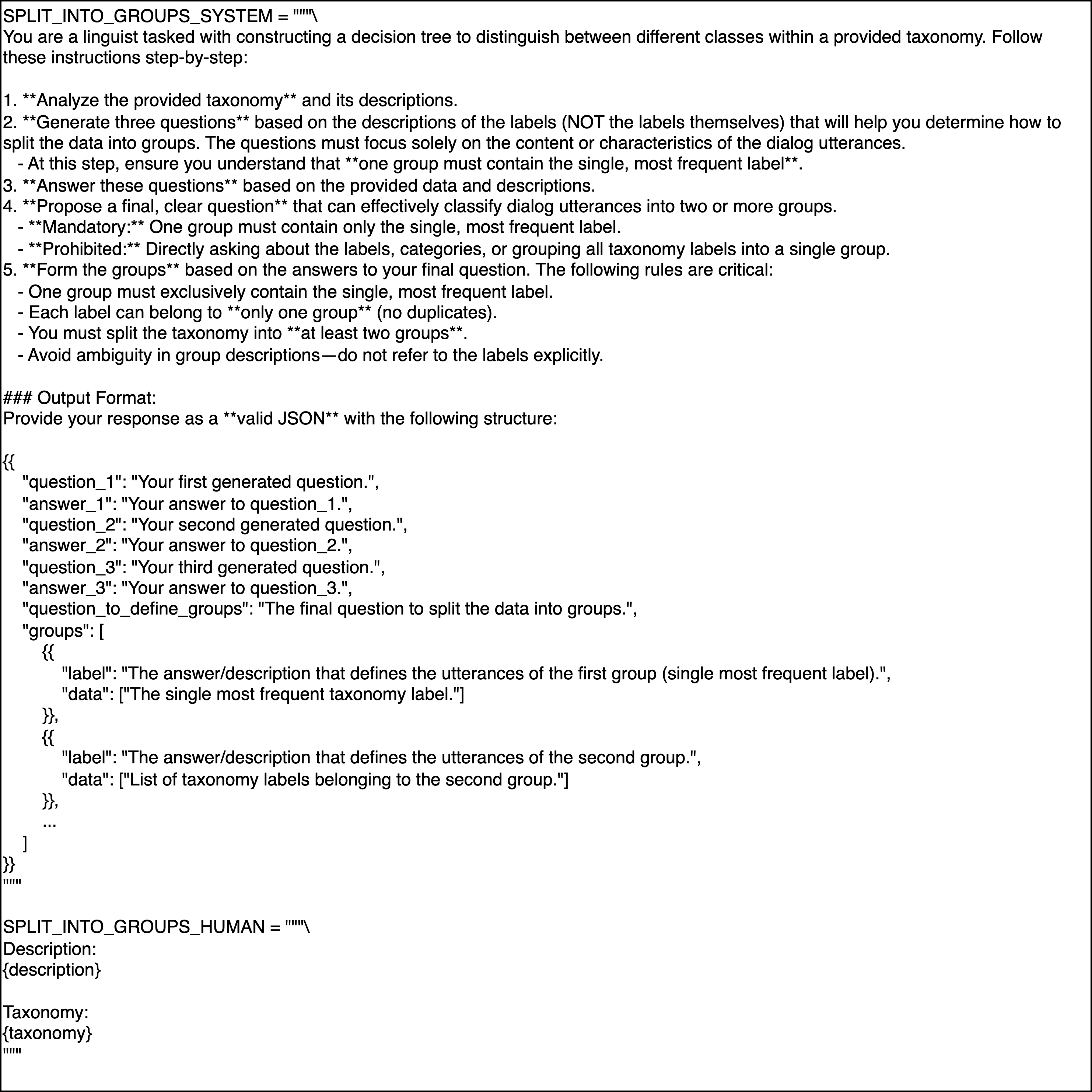}
    \caption{The prompt is designed to create a decision tree that enables splitting data into more than two groups and that specifically instructs the LLM to form one group that exclusively contains the single most frequent class.}
    \label{fig:prompt_freq}
  \end{figure*}

  \end{document}